\documentclass{article} 
\usepackage{iclr2024_conference,times}

\usepackage{amsmath,amsfonts,bm}

\def\eqref#1{equation~\ref{#1}}

\def\1{\bm{1}}

\DeclareMathAlphabet{\mathsfit}{\encodingdefault}{\sfdefault}{m}{sl}
\SetMathAlphabet{\mathsfit}{bold}{\encodingdefault}{\sfdefault}{bx}{n}

\usepackage{hyperref}
\usepackage{url}
\usepackage{graphicx}
\usepackage{caption, subcaption}

\title{Enabling Clean Energy Resilience with\\Machine Learning-Empowered\\Underground Hydrogen Storage}

\iclrfinalcopy

\author{Alvaro Carbonero, Shaowen Mao \& Mohamed Mehana \\
Los Alamos National Laboratory\\
Los Alamos, New Mexico, USA \\
\texttt{\{alvaro.carbonero, shaowen.mao, mzm\}@lanl.gov} \\
}

\begin{document}

\maketitle

\begin{abstract}
To address the urgent challenge of climate change, there is a critical need to transition away from fossil fuels towards sustainable energy systems, with renewable energy sources playing a pivotal role. However, the inherent variability of renewable energy, without effective storage solutions, often leads to imbalances between energy supply and demand. Underground Hydrogen Storage (UHS) emerges as a promising long-term storage solution to bridge this gap, yet its widespread implementation is impeded by the high computational costs associated with high fidelity UHS simulations. This paper introduces UHS from a data-driven perspective and outlines a roadmap for integrating machine learning into UHS, thereby facilitating the large-scale deployment of UHS.
\end{abstract}

\section{Introduction}

Renewable energy, a key player in combating climate change, is gaining increasing global adoption \cite{Energy20202010EuropeanCommission,Unitednations2015AgreementP}. In 2022, renewables contributed 13.1$\%$ to the US's primary energy consumption \cite{U.S.EnergyInformationAdministrationEIA2022MonthlyReview} and 21.5$\%$ to its utility-scale electricity generation. Furthermore, by 2024, countries like Spain, Germany, and Ireland are expected to generate over 40$\%$ of their annual electricity from wind and solar photovoltaics \cite{U.S.EnergyInformationAdministrationEIA2023Renewable2024}. Despite these strides, renewable energy sources face challenges due to unpredictable variations in atmospheric conditions and geographic limitations, leading to fluctuations in energy supply. 

Underground hydrogen (H$_2$) storage (UHS) is emerging as a vital technology for mitigating the intermittency of renewable energy sources \cite{zivar2021underground, tarkowski2022towards}. When there's a surplus in renewable energy production,  excess electricity is converted into H$_2$ and stored in geological formations. During times of high demand or low renewable energy generation, the stored H$_2$ is then retrieved and utilized to meet energy needs. UHS, unlike batteries typically used for short-term storage \cite{raad2022hydrogen}, is capable of storing significant amounts of energy over longer periods. This capability is crucial for counteracting seasonal energy fluctuations and maintaining a consistent energy supply year-round. 

UHS shares similarities with other subsurface applications such as hydrocarbon development \cite{christie2001tenth} and geological carbon sequestration (GCS) \cite{moridis2023practical,wen2023real}. However, it is distinguished by more complex operational conditions. In hydrocarbon development, the process is typically focused on extraction, while GCS is solely about injection. In contrast, UHS operates on a cyclic basis, incorporating both injection and extraction stages. This complexity in operational conditions introduces greater uncertainty in H$_2$ storage performance. Traditionally, the prediction of UHS performance relies on high-fidelity physics-based reservoir simulations \cite{lysyy2021seasonal,feldmann2016numerical,hogeweg2022benchmark,okoroafor2023intercomparison}. These simulations accurately predict the H$_2$ movement and pressure changes in geological formations during UHS operations. However, they are extremely computationally intensive, thus delaying the pace of large-scale UHS deployment. 
To accelerate UHS prediction, turning to surrogate modeling via machine learning (ML) offers a promising strategy. 

Currently, there is a knowledge gap in the literature when it comes to developing ML surrogate models for UHS--with only two articles on the subject up to our knowledge \cite{mao2024efficient, chen2024integrating}. This gap is attributed not only to the fact that UHS is an emerging field but also to the inherent complexity of UHS operational conditions. It is important to note that the knowledge and techniques used in surrogate modeling for GCS do not directly apply to UHS due to its distinct and intricate operational dynamics. In this study, we outline the unique challenges of applying ML to UHS and propose ideas for developing accurate and efficient UHS ML models, supporting risk assessment and storage optimization of future UHS operations.

\section{Data generation}
\label{sec:data_generation}

UHS simulations are conducted by solving partial differential equations (PDEs) the details of which can be found in Appendix \ref{app:pde}. While data for these simulations can be generated using reservoir simulators with specific input configurations, there is a hope that future work will make such datasets publicly available. The computational cost of these simulations increases significantly with refined grids or an increased number of components in the system. For our preliminary studies, we utilized tNavigator\cite{RockFlowDynamics_2023} to perform 1000 2D UHS simulations in depleted gas reservoirs, varying porosity and permeability heterogeneity. Each simulation includes ten annual cycles of six-month H$_2$ injection followed by six months of withdrawal. These simulations were parallelized across four 64-core CPU servers, taking approximately five days to complete. An example is depicted in Figure \ref{fig:data_example}. See Appendix \ref{app:dataset} for more details on the data produced, and Figure \ref{fig:fluvial_example} for a simulation conducted in a different type of geological formation. It should be noted that data generation of more realistic systems, such as ones that are not 2D, that will require significantly more computing power.

\begin{figure}
    \centering
    \includegraphics[scale=0.36]{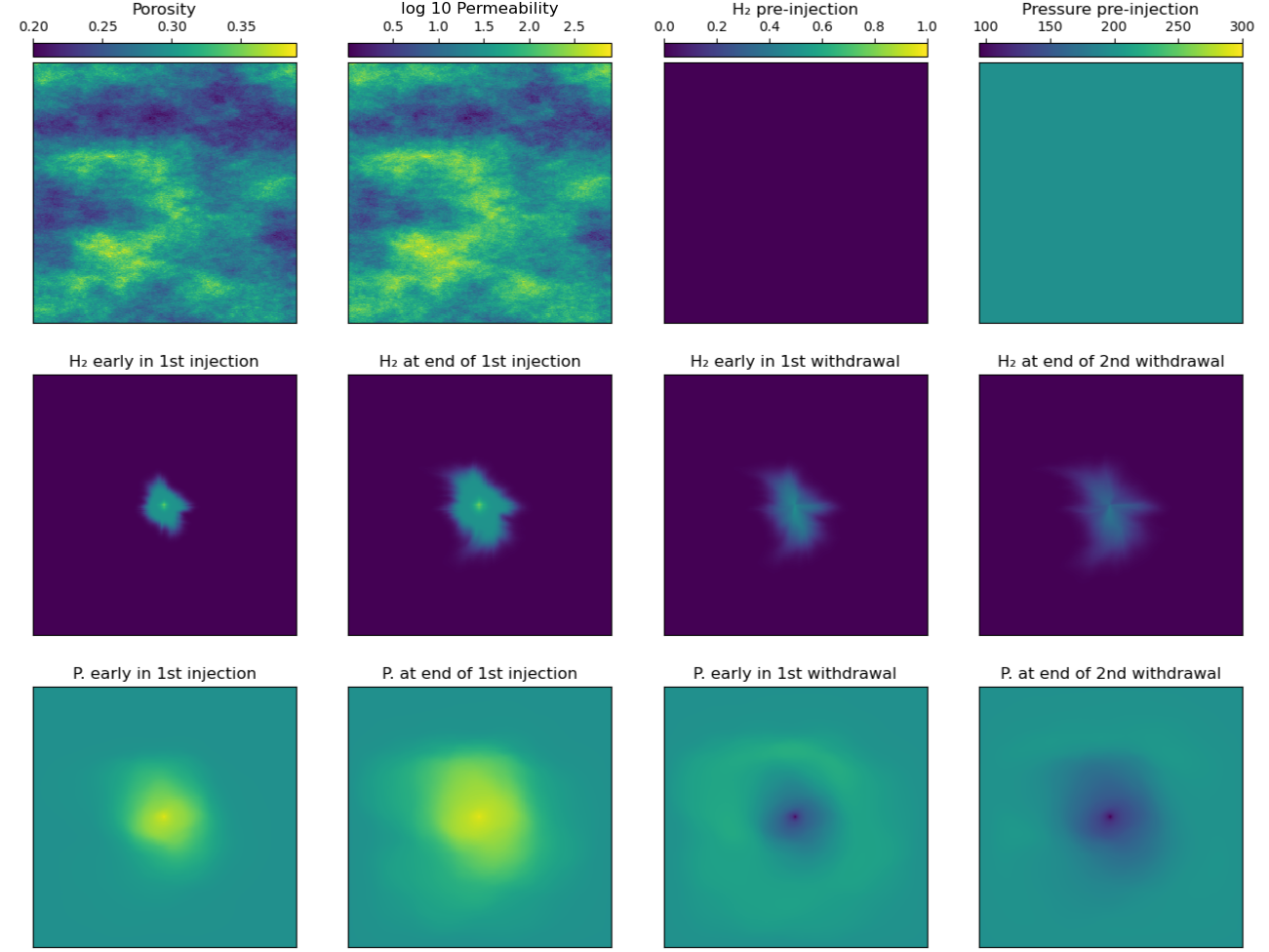}
    \caption{Temporal evolution of the spatial distribution of H$_2$ saturation and reservoir pressure in a 2D UHS simulation. H$_2$ is injected and withdrawn from a central well in a depleted gas reservoir over ten annual cycles comprising a 6-month injection stage followed by a 6-month withdrawal stage. The initial two subfigures (top row) depict the heterogeneous porosity and permeability of the geological formation. Subsequent figures show the H$_2$ saturation and pressure distributions at various time points during UHS operations. `Early' of a stage refers to two months after its onset, and `end' of a stage to six months. Porosity is dimensionless; permeability is in millidarcys ($10^{-15} m^2$); H$_2$ saturation is dimensionless; pressure is in bars ($10^5$ Pa).}
    \label{fig:data_example}
\end{figure}

\section{Roadmap towards ML for UHS}
\label{sec:challenges}

We outline the required capabilities for a UHS ML model, reference GCS research with potential applicability to UHS, and identify their limitations due to the complexity of UHS operations.

\subsection{Model parameters}
\label{subsec:minimum}

A UHS ML model receives many uncertain input parameters, and predicts many output parameters. The \textbf{uncertain input parameters} for UHS models include geological and operational parameters. Geological parameters typically involve reservoir depth, thickness, permeability, porosity, and geothermal gradient. Operational parameters commonly cover the injection-withdrawal scheme, injection pressure, production pressure, the number of storage cycles, and cushion gas type and volume (the cushion gas is injected before H$_2$ cycling to provide pressure support for H$_2$ recovery). The \textbf{output parameters} crucial for UHS projects fall into two categories. The first involves spatial distributions of key variables, including H$_2$ saturation, reservoir pressure, and surface deformation. The second includes scalar metrics for evaluating storage performance, such as H$_2$ recovery factor, purity of produced H$_2$, gas-water ratio, and well injectivity. 


\subsection{Translating GCS methods to UHS}

There is abundant research that applies ML to GCS. The methods developed can be divided into deep learning architectures, such as graph neural networks \cite{ju2023learning}, and neural operators, such as nested Fourier Neural Operators (FNO) \cite{wen2023real}.
Since GCS and UHS simulations involve the same set of PDEs (with different boundary conditions at the well), a good first stride towards ML for UHS is translating these frameworks to the UHS setting.
Nonetheless, we foresee three challenges when translating GCS models to UHS.

\subsection{Challenges to overcome}

The first challenge involves \textbf{developing successful auto-regressive models}. An auto-regressive model uses its previous outputs as inputs to generate future predictions. This contrasts with the static models commonly found in GCS literature \cite{wen2022u, diab2023u}, which directly predict a specific point in time by taking the desired time as an input parameter. 
Auto-regressive models offer advantages, including the ability to extrapolate over time with reasonable accuracy, something static models cannot do. 
By leveraging time extrapolation and decreasing the simulated time, models can be trained over a wider range of uncertain input parameters
Time extrapolation can allow for training over a wider range of uncertain input parameters by decreasing simulation time. Additionally, they can leverage monitoring data from ongoing UHS projects to enhance future predictions without model retraining or tuning, helping with subsequent operational optimization.
However, there is a common concern for the performance of auto-regressive models due to their potential to accumulate errors over time, a problem that could be compounded by the complex operational dynamics of UHS.
Nonetheless, we provide preliminary results in Appendix \ref{app:comparison} that, despite error accumulation, auto-regressive models perform comparatively well to static ones in the UHS setting.


The second challenge is \textbf{modifying GCS architectures to predict scalar values}. Unlike GCS, predicting the temporal evolution of the spatial distribution of H$_2$ saturation, reservoir pressure, and surface deformation is insufficient for UHS projects. 
To evaluate the risks and optimize the storage operations, models also need to predict scalar values that are critical UHS performance metrics, such as those mentioned in Section \ref{subsec:minimum}. 
We are assessing the possibility of tweaking the decoders of GCS models to predict both spatial distributions and scalar values.

The third challenge lies in \textbf{creating models that produce real-time high-resolution UHS predictions}. Although GCS methods exist, such as nested FNOs, UHS may demand high resolution in larger spaces than GCS. Cushion gasses increase the size of the gas plume, cyclic injection schemes complicate the plume's behavior, and deployment in geological formations with preferential paths increase plume migration. Dealing with these factors may demand high-resolution in larger areas than in GCS operations. We elaborate on this in Appendix \ref{app:grid_refinement}.

\subsubsection*{Acknowledgments}

This work was supported by Los Alamos National Laboratory Technology Evaluation \& Demonstration funds and Laboratory Directed Research and Development (LDRD) program. This work was supported by the Environmental Management-Minority Serving Institution Partnership Program (EM-MSIPP) Graduate Fellowship. This research used resources provided by the Los Alamos National Laboratory Institutional Computing Program, which is supported by the U.S. Department of Energy National Nuclear Security Administration under Contract No. 89233218CNA000001.

\bibliography{iclr2024_conference}
\bibliographystyle{iclr2024_conference}

\newpage

\appendix
\section{The governing equations of state of an UHS simulation}
\label{app:pde}

The mass balance equations dictate the behavior of H$_2$, water and the cushion gas in the subsurface and are explicitly stated below.

\begin{align*}
    \frac{\partial}{\partial t} \left(\sum_{\beta \in \{\text{A, G} \}}\phi S_{_{\beta}}\rho_{_{\beta}}X_{_{\beta}}^{k} \right) - \nabla \cdot \left( \sum_{\beta \in \{\text{A, G} \}} \rho_{_{\beta}}X_{_{\beta}}^k F_{_{\beta}} \right) = q^{k} & \hfill & \text{for }k=\text{H$_2$, water, C}
\end{align*}
where C is the cushion gas being used, A and G are the aqueous and gaseous phases respectively, $\phi$ is porosity, $S_{\beta}$ is the saturation of phase $\beta$, $\rho_{\beta}$ is the density of phase $\beta$, $X_{\beta}^k$ is the mass fraction of component $k$ in phase $\beta$, $q^k$ is the source/sink term of component $k$, and $F_{\beta}$ is the advective mass flux of phase $\beta$. The latter value is obtained from Darcy's law:
\[
F_{\beta} = -\frac{K\cdot K_{r\beta}}{\mu_{\beta}}\cdot \left( \nabla P_{\beta} - \rho_{\beta} G)\right)
\]
where $K$ is the rock's intrinsic permeability, $K_{r\beta}$ the relative permeability of phase $\beta$, $\mu_{\beta}$ is the viscosity of phase $\beta$, $P_{\beta}$ is the pressure of phase, and $G$ is the gravitation accelerator vector.


\section{Comparing static and auto-regressive models using U-net}
\label{app:comparison}

In this appendix, we compare the performance of static and auto-regressive models using U-net, a popular image-to-image model. Our objective is to show brief experimental data that auto-regressive models do not accumulate a significant amount of error and that auto-regressive models can extrapolate in time successfully. We trained four U-net models to predict H$_2$ saturation or pressure following the static or the auto-regressive frameworks.  

\subsection{Dataset}
\label{app:dataset}

The dataset consists of the simplest setup for UHS. There are 1000 simulations, each with a unique porosity and permeability map with fixed operational parameters. The task is to predict the progression of the H$_2$ plume and pressure. Every simulation goes up to 10 years with a time output every 2 months. Every 6 months, the simulation changes between injection and withdrawal. Each simulation is 2D in a 7680m x 7680m grid with the well always located in the middle. To approximate a 2D behavior, the simulator gives the reservoir a thickness of 100m. Even though the simulation was produced with 256 resolution, the dataset's resolution was downgraded to 64 resolution to speed up training in these preliminary results.

The dataset is split into 4 parts as follows:
\begin{itemize}
    \item Training dataset: this dataset consists of 700 simulations with simulation time going up to 7 years.
    \item Validation set 1 (time extrapolation): this dataset consists of the 700 simulations used in the training set and using the leftover 3 years of simulation time.
    \item Validation set 2 (geological extrapolation): this dataset consists of 150 simulations not used in the training set  with simulation time going up to 7 years.
    \item Validation set 3 (geological and time extrapolation): this dataset consists of the 150 simulations used in validation set 2 and using the leftover 3 years of simulation time.
\end{itemize}

The leftover 150 simulations of the 1000 total simulations are reserved as a testset, but we will not do experimentation on them in this paper. \textbf{For performance comparisons between static and auto-regressive models, we only use Validation set 2}.

The input parameters are indicated in Table \ref{tab:inputs} and as follows. \textbf{Distance to well:} Since the well is always located in the middle, the distance to well channel is transformed so that values near the well are highest, and then normalized to values between 0 and 1. Our experiments show that this channel is very important for good predictions.
\textbf{Cycle stage:} The channel is a broadcast of 1 if the time step being predicted is of injection, and -1 otherwise. Our experiments show that this channel is very important for good predictions.

\begin{table}[]
\centering
\caption{The types of input the models take as well as their unit and transformation.}
\label{tab:inputs}
\begin{tabular}{lllll}
\textbf{Type of input}       & \textbf{Unit}  & \textbf{Transformation}     & \textbf{In static model?} & \begin{tabular}[c]{@{}l@{}}\textbf{In auto-reg. model?}\\ \end{tabular} \\ \hline \\
Porosity            & /      & Mean 0, variance 1 & Yes              & Yes                                                                 \\ 
Permeability        &   mD    & Mean 0, variance 1 & Yes              & Yes                                                                 \\ 
H$_2$ saturation & /    & None               & No               & \begin{tabular}[c]{@{}l@{}}Yes (if part\\ of output)\end{tabular}   \\ 
Pressure            & $10^{5}$ Pa   & Mean 0, variance 1 & No               & \begin{tabular}[c]{@{}l@{}}Yes (if part of\\ output)\end{tabular}   \\ 
Distance to well    & m  & See above          & Yes              & Yes                                                                 \\ 
Cycle indicator     & /  & See above          & Yes              & Yes                                                                 \\ 
Time                & 2 months & Divided by 60      & Yes              & No                                                                  \\ 
\\
\end{tabular}
\end{table}

\subsection{U-net architectures}

All four models share the same architecture. Each model has three down/upsampling layers, with initial embedding size of 64 and ReLU activation functions. The models for H$_2$ saturation have a final sigmoid activation layer while pressure models have a final tanhshrink activation layer. The error function is mean absolute error (MAE) for all of them.

\subsection{Training and results}

Each model is trained with a batch size of 64. The initial learning rate is 0.0001, halved every 25 epochs. Training is set for 200 epochs, but training is halted if 10 consecutive epochs pass with no improvements in training error. The model gets saved whenever a new average validation error is achieved, which is calculated every 10 epochs. This helps guarantee that the output model does not overfit. In static models, only Validation set 2 is used to decide whether to save the model as the other validation sets test extrapolation in time. We noticed in our experiments that static models are prone to overfitting, so we added L2 regularization. The models were trained on an A100 GPU with 40GB of memory.

See Table \ref{tab:unetresults} for the results. Of note is that auto-regressive models perform significantly better when they are fed perfect information, which is the case during training. Auto-regressive models outperform static models in validation error by $86.1\%$ and $71.0\%$ in H$_2$ saturation and pressure respectively. Nonetheless, as we will see, this gap in performance changes in long roll-outs.

\begin{table}[t]
\caption{Model training results.}
\label{tab:unetresults}
\centering
\begin{tabular}{lll}
\textbf{Model}                       & \textbf{Training error} & \textbf{Validation error} \\ \hline \\
H$_2$ saturation, auto-regressive & 0.000237                & 0.000373                  \\
H$_2$ saturation, static          & 0.00216                 & 0.00269                   \\
Pressure, auto-regressive            & 0.0224                  & 0.0544                    \\
Pressure, static                     & 0.00873                 & 0.0158                    \\
\end{tabular}
\end{table}

\subsection{Accumulation of error in auto-regressive models}

\begin{figure}
    \centering
    \begin{subfigure}[b]{0.49\textwidth}
        \centering
        \includegraphics[width=2.7in]{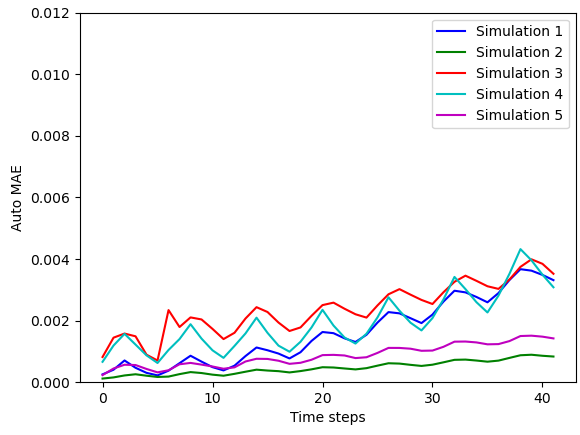}
        \caption{H$_2$ auto-regressive's MAE against time}
        \label{fig:hgasauto}
    \end{subfigure}
    \hfill
    \begin{subfigure}[b]{0.49\textwidth}
        \centering
        \includegraphics[width=2.7in]{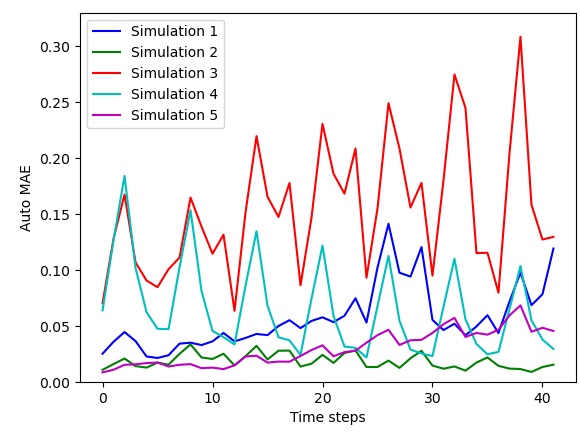}
        \caption{Pressure auto-regressive MAE against time}
        \label{fig:pressureauto}
    \end{subfigure}
    \begin{subfigure}[b]{0.49\textwidth}
        \centering
        \includegraphics[width=2.7in]{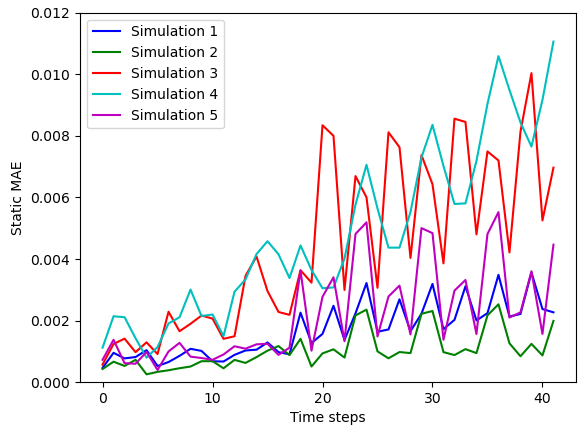}
        \caption{H$_2$ static MAE against time}
        \label{fig:hgasstatic}
    \end{subfigure}
    \hfill
    \begin{subfigure}[b]{0.49\textwidth}
        \centering
        \includegraphics[width=2.7in]{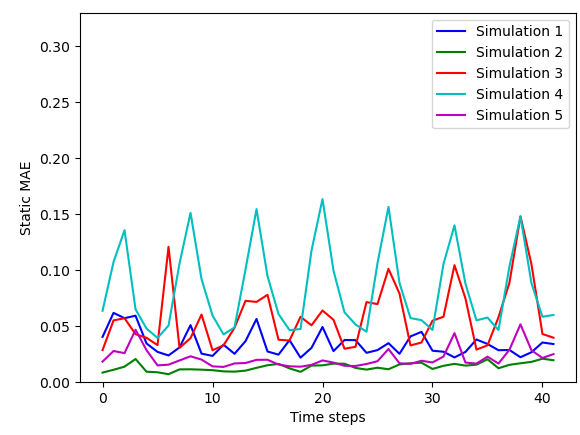}
        \caption{Pressure static MAE against time}
        \label{fig:pressurestatic}
    \end{subfigure}
    \begin{subfigure}[b]{0.49\textwidth}
        \centering
        \includegraphics[width=2.7in]{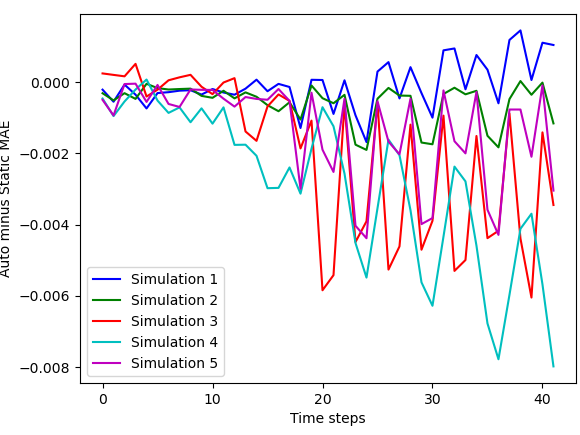}
        \caption{H$_2$, auto minus static MAE against time}
        \label{fig:hgasautostatic}
    \end{subfigure}
    \hfill
    \begin{subfigure}[b]{0.49\textwidth}
        \centering
        \includegraphics[width=2.7in]{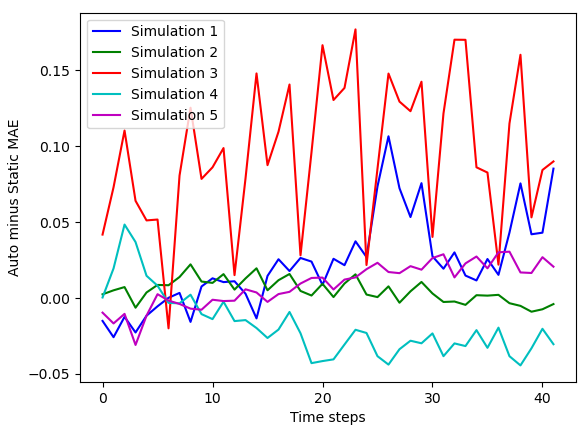}
        \caption{Pressure, auto minus static MAE against time}
        \label{fig:pressureautostatic}
    \end{subfigure}
        \caption{Comparison between static and auto-regressive models. Five simulations were randomly selected from the validation set for the comparison. There are 42 time steps in each plot, where one time step is equivalent to 2 months. Figures \ref{fig:hgasauto}, \ref{fig:hgasstatic}, \ref{fig:hgasautostatic} concern H$_2$ saturation models, while Figures \ref{fig:pressureauto}, \ref{fig:pressurestatic}, \ref{fig:pressureautostatic} concern pressure models. Figures \ref{fig:hgasautostatic}, \ref{fig:pressureautostatic} plot time steps against the MAE of the auto-regressive model minus the MAE of the static model.}
        \label{fig:autovsstatic}
\end{figure}

To contrast error accumulation in auto-regressive models against static model errors, we will show three different types of plots. The first two types of plots are the progression of the models against time. The last type shows the difference in MAE of auto-regressive models and static models against time. We recognize that this comparison cannot definitively settle whether auto-regressive models can perform as well as static models despite error accumulation. Nonetheless, we believe that our results are promising and point to the need for further studies to create architectures or training schemes tailored to the auto-regressive setting. For this proposal, we will do this only for 5 randomly chosen simulations from the validation set. Nonetheless, we hope to do more thorough studies in the future. See Figure \ref{fig:autovsstatic} for the results.

\textbf{Auto-regressive versus static: } In Figures \ref{fig:hgasautostatic}, \ref{fig:pressureautostatic} we compare the numerical difference between auto-regressive models and static models directly by plotting time steps against the MAE of auto-regressive models minus the MAE of static models. If the values are negative, then the auto-regressive model had better performance than the static model. Thus, it is surprising to see an uneven result. Figure \ref{fig:hgasautostatic} shows that H$_2$ auto-regressive models outperformed static models, while \ref{fig:pressureautostatic} shows the opposite for pressure. Surprisingly, the difference for $H_2$ models widens as time steps increases despite the accumulation of error in auto-regressive models. 

\textbf{H$_2$ models: } Figure \ref{fig:hgasstatic} hints that as time steps increases, the predictions get harder and harder to do for H$_2$ static models. This might be because as the plume increases, the task to predict the intricacies of the plume get harder. Moreover, as time progresses, the hydrogen plume settles into progressively more static positions, thus making it harder for a static model to capture the fine differences between different time steps. This difficulty does not seem to arise in the auto-regressive model which have a weak upward trend when compared to the upward trend of the auto-regressive model.

\textbf{Pressure models: } Figure \ref{fig:pressurestatic} shows no upward trend on the pressure static model, but surprisingly, Figure \ref{fig:pressureauto} shows only a weak upward trend in the pressure auto-regressive model. This hints at the fact that accumulation of error is not significant enough to make the pressure static model significantly outperform the pressure auto-regressive model.

\textbf{Conclusions: } This brief empirical result motivates the need for further study on the true difference in performance between auto-regressive models and static models in the setting of UHS. Static models seem to struggle when predicting large time steps. While auto-regressive models have much better training errors, it seems that this difference in performance disappears in long-term predictions probably due to the accumulation of error. Thus, there seems to be two possible directions for further study: either to make static models that do well in large time steps, or to make auto-regressive architectures or training schemes that accumulate less error. The latter has the additional operational benefits we mentioned in Section \ref{sec:challenges}.

\subsection{Extrapolation over time in auto-regressive models}

\begin{figure}
    \centering
    \begin{subfigure}[b]{0.49\textwidth}
        \centering
        \includegraphics[width=2.7in]{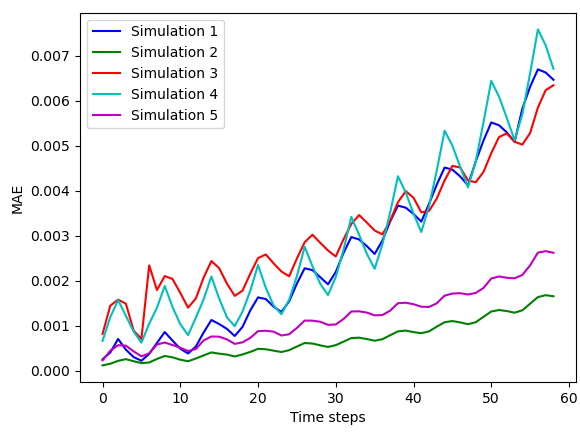}
        \caption{H$_2$ auto-regressive}
        \label{fig:hgastime}
    \end{subfigure}
    \hfill
    \begin{subfigure}[b]{0.49\textwidth}
        \centering
        \includegraphics[width=2.7in]{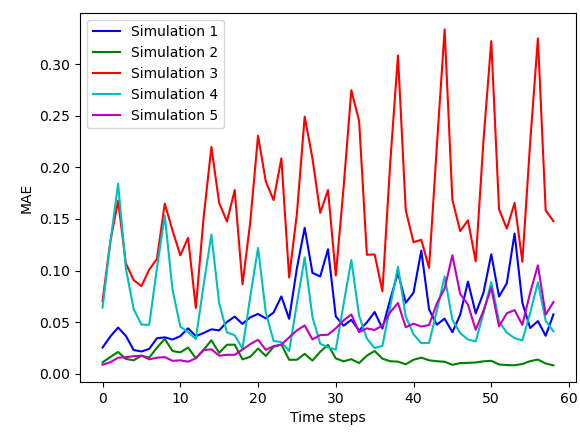}
        \caption{Pressure auto-regressive}
        \label{fig:pressuretime}
    \end{subfigure}
        \caption{Demonstration of extrapolation in time of auto-regressive models. Figure \ref{fig:hgastime} demonstrates this for the H$_2$ model, while Figure \ref{fig:pressuretime} demonstrates it for the pressure model.}
        \label{fig:timeextrapolation}
\end{figure}

In this section, we provide brief empirical evidence that auto-regressive models can extrapolate in time successfully. Figure \ref{fig:timeextrapolation} shows the error of auto-regressive models up to time step 60. Since the last time step in the training set is 42, this shows that the models are able to do reasonably well when extrapolating in time. It is worth noting that Figure \ref{fig:hgastime} seems to have a persistent upward trend, while this trend seems to be absent in Figure \ref{fig:pressuretime}. This behavior hints at the fact that H$_2$ models have significantly different behaviors than pressure models. Their difference in unit magnitude, where H$_2$ values range from 0 to 1 and pressure values range from negative to positive values, might provide an explanation. 

\section{High-resolution Modeling for H$_2$ Plume Dynamics}
\label{app:grid_refinement}

\begin{figure}
    \centering
    \includegraphics[scale=0.40]{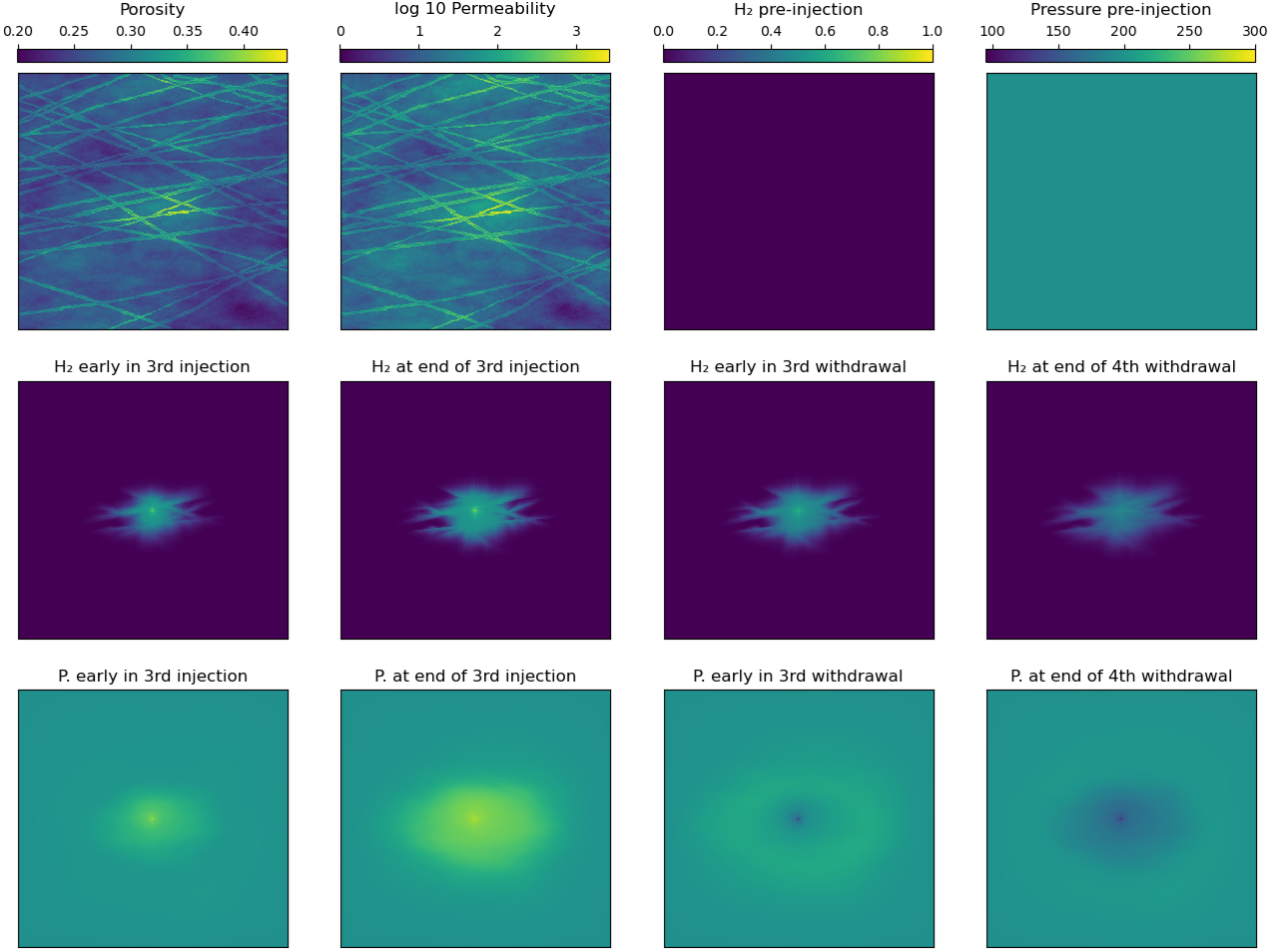}
    \caption{A 2D UHS simulation that captures the H$_2$ plume and pressure as time progresses. H$_2$ is injected and withdrawn in cycles of 6 months each, where the simulator further subdivides each 6 month stage into 3 parts. By start of a stage, we mean 2 months after it has started as this is the earliest when the simulator saves data. By end of a stage, we mean 6 months after it has started.    The simulation is carried out in a different type of mapping than in Figure \ref{fig:data_example}. This figure illustrates the preferential paths that the hydrogen plume takes, making the plume go far from the well. Coarsening the image in the outer parts of the plume would delete the preferential paths of the plume. This can potentially obstruct the model learning the true behavior of the plume in these kinds of geological formations.}
    \label{fig:fluvial_example}
\end{figure}

Geological formations are commonly characterized by two types of heterogeneity: fluvial and Gaussian fields. In Gaussian fields, properties such as porosity and permeability are distributed based on a normal (Gaussian) distribution. This distribution leads to gradual and predictable variations in these properties throughout the formation. Conversely, fluvial fields derive their heterogeneity from the sediment deposition processes of rivers and streams. This results in intricate patterns of sand bodies, silt, and clay, exhibiting a wide range of properties that reflect the dynamic interplay of water flow and sediment transport. Due to the complex nature of these deposits, fluvial fields often feature preferential flow paths with high porosity and permeability. An illustration of fluvial field heterogeneity is shown Figure \ref{fig:fluvial_example}.

Compared to CO$_2$, H$_2$ exhibits lower viscosity and higher diffusivity in subsurface environments. In fluvial fields characterized by highly permeable channels, the footprint of an H$_2$ plume can expand significantly, resulting in the gas plume front extending far from the injection well. The preliminary injection of cushion gas prior to H$_2$ cycling can further enlarge the gas plume. To precisely model the plume's shape, a high-resolution computational grid is necessary across a broad section of the simulation domain. The approach of applying drastic local grid refinement near the well, as commonly done in previous geological carbon storage (GCS) projects, may not be adequate for UHS modeling, due to the distinct behavior of H$_2$ in the subsurface.

\end{document}